\documentclass[letterpaper]{article} 
\usepackage[preprint]{aaai2027}  

\usepackage[hyphens]{url}  
\usepackage{graphicx} 
\usepackage{natbib}  
\usepackage{caption} 
\usepackage{algorithm}
\usepackage{algorithmic}
\usepackage{amsmath}
\usepackage{amssymb}
\usepackage{booktabs}  
\usepackage{multirow}  
\usepackage[table]{xcolor} 
\usepackage{siunitx}

\DeclareMathOperator{\Clip}{Clip}
\DeclareMathOperator{\Round}{Round}
\usepackage{newfloat}
\usepackage{listings}
\DeclareCaptionStyle{ruled}{labelfont=normalfont,labelsep=colon,strut=off} 
\floatstyle{ruled}
\newfloat{listing}{tb}{lst}{}
\floatname{listing}{Listing}

\usepackage{booktabs}

\title{EM-SNN: Efficiently Modulated Spiking Neural Network for \\ Remote Sensing Image Dehazing}

\author{
    Jie Shao\textsuperscript{\rm 1},
    Jiaqi Ma\textsuperscript{\rm 2},
    Wenwen Min\textsuperscript{\rm 3},
    Beihang Song\textsuperscript{\rm 4},
    Ning Chen\textsuperscript{\rm 1},
    Youfa Liu\textsuperscript{\rm 5},
    Jun Wan\textsuperscript{\rm 1}\corresponding
}

\affiliations{
    \textsuperscript{\rm 1}Zhongnan University of Economics and Law, Wuhan, China\\
    \textsuperscript{\rm 2}Mohamed bin Zayed University of Artificial Intelligence, Abu Dhabi, UAE\\
    \textsuperscript{\rm 3}Yunnan University, Kunming, China\\
    \textsuperscript{\rm 4}National Institute of Natural Hazards, Ministry of Emergency Management of China, Beijing, China\\
    \textsuperscript{\rm 5}Wuhan University, Wuhan, China
}

\begin{document}

\maketitle

\begin{abstract}
Although spiking neural networks (SNNs) provide an energy-efficient alternative to artificial neural networks (ANNs), their application to remote sensing image dehazing remains limited. A key challenge arises from the coupling between haze-induced high-frequency attenuation and discrete spike thresholding. This interaction suppresses weak responses and fundamentally limits the recovery of edges, textures, and fine details in spiking dehazing models. To address this challenge, we propose the \textbf{E}fficiently \textbf{M}odulated \textbf{S}piking \textbf{N}eural \textbf{N}etwork (EM-SNN), a dedicated spiking framework tailored to remote sensing image dehazing. EM-SNN integrates a statistics-driven Threshold-Modulated Leaky Integrate-and-Fire (TM-LIF) neuron to adaptively compensate for haze-induced contrast compression, together with a Spike Sobel Modulation (SSM) module that enhances structural cues and reduces depth-wise attenuation during spiking feature propagation. By jointly modulating activation scales and structural representations, EM-SNN improves dehazing performance while preserving the inherent event-driven sparsity of SNNs. Experiments on HRSD, RICE, RRSHID, and SateHaze1K demonstrate that EM-SNN achieves competitive dehazing performance while consuming only one quarter of the energy of the strong ANN baseline SFRDP-Net.
\end{abstract}

\section{Introduction}
Remote sensing images provide detailed ground scene information and are widely used in numerous applications,  including agricultural production~\cite{mladenova2019evaluating}, environmental protection~\cite{jiang2022fusion}, and disaster monitoring~\cite{zhang2002flood}. Haze and atmospheric scattering degrade image quality by reducing contrast and obscuring structural details. This degradation is highly frequency-dependent, with high-frequency components (e.g., edges and textures) being severely attenuated while low-frequency components are largely preserved, resulting in over-smoothed images with lost details. Although deep learning models can recover such structures,  they often come with a trade-off of increased computational overhead and growing storage costs, reducing their efficacy in practical scenarios~\cite{sun2025spatial,hu2025clusir}.

\begin{figure}
  \includegraphics[width=1.0\linewidth]{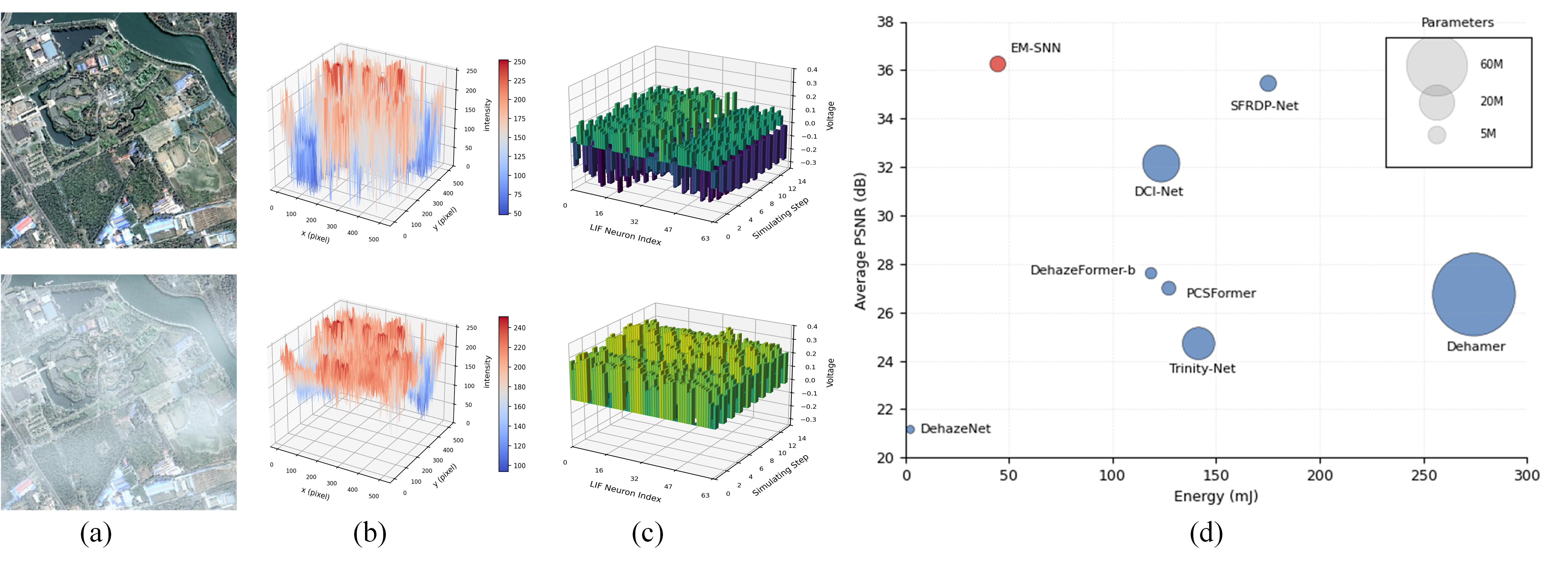}
  \vspace{-1.0em}
  \caption{Visualization of haze-induced degradation and spiking responses.
    (a) Remote-sensing patch under clear (top) and hazy (bottom) conditions.
    (b) 3D intensity surface plots of the patches.
    (c) LIF membrane potentials over neuron index and simulation step.
    (d) PSNR versus energy consumption on RICE-2; bubble size indicates parameter.}
  \label{fig:teaser}
  \vspace{-2.0em}
\end{figure}

More recently, brain-inspired SNNs, as the third generation of neural networks, have emerged as a promising alternative for energy-efficient intelligence. Unlike conventional ANNs that encode features as continuous values, SNNs communicate with discrete spikes and propagate information through event-driven neuronal dynamics, facilitating deployment on neuromorphic hardware with mW-scale power budgets~\cite{yao2024spike}. These properties make SNNs particularly attractive for remote sensing dehazing under constrained onboard computation and power budgets.

However, extending SNNs to remote sensing image dehazing remains challenging. Remote sensing scenes contain abundant small objects, thin boundaries, and fine-grained textures, whose feature responses are particularly vulnerable to haze-induced attenuation. Haze compresses local contrast and shifts these weak structural responses into a low-activation regime, resulting in compounded information loss when they are subsequently encoded by discrete spikes. Under fixed firing thresholds, the resulting amplitude--threshold mismatch causes many weak edge and texture activations to remain below the firing boundary, making them under-represented or even discarded in the spike domain. Moreover, repeated spike quantization suppresses subtle activation variations, while membrane leakage makes weak responses more difficult to accumulate across time. These effects progressively attenuate high-frequency structural information as network depth increases. As shown in Fig.~\ref{fig:teaser}(a--c), compared with clear inputs, hazy inputs produce lower membrane-potential responses and a larger proportion of sub-threshold activations, particularly around weak edges and textures.

Motivated by this analysis, we propose EM-SNN, an energy-efficient spike-driven framework for remote sensing image dehazing, composed of a Threshold-Modulated Leaky Integrate-and-Fire (TM-LIF) neuron and a Spike Sobel Modulation (SSM) module. 
The TM-LIF neuron calibrates spike quantization via a statistics-driven, channel-wise adaptive threshold, improving robustness under haze-induced contrast compression and stabilizing feature scales across depth. 
The SSM module uses Sobel gradients as a lightweight structural prior to perform channel and spatial modulation, compensating for the depth-wise attenuation of weak structural details during spiking feature propagation. Together, TM-LIF and SSM provide coupled scale-and-structure modulation for high-quality remote sensing image dehazing. As shown in Fig.~\ref{fig:teaser}(d), the proposed EM-SNN significantly reduces energy consumption while maintaining competitive PSNR compared with existing approaches. Our contributions are as follows:
\begin{itemize}
\item A TM-LIF neuron is proposed to calibrate spike quantization via a statistics-driven, channel-wise adaptive threshold for stabilizing multispike representations and mitigating information loss caused by haze-induced contrast compression.

\item  A deployment-friendly and energy-aware SSM module is designed to leverage Sobel-derived structural priors to compensate for the depth-wise attenuation of weak structural details in deep spiking feature propagation.

\item  By collaborating TM-LIF neuron and SSM module within the proposed EM-SNN, activation scales and structural representations are more efficiently modulated, achieving competitive dehazing performance with significantly reduced energy consumption.
\end{itemize}

\begin{figure*}[t]
  \centering
  \includegraphics[width=0.92\linewidth]{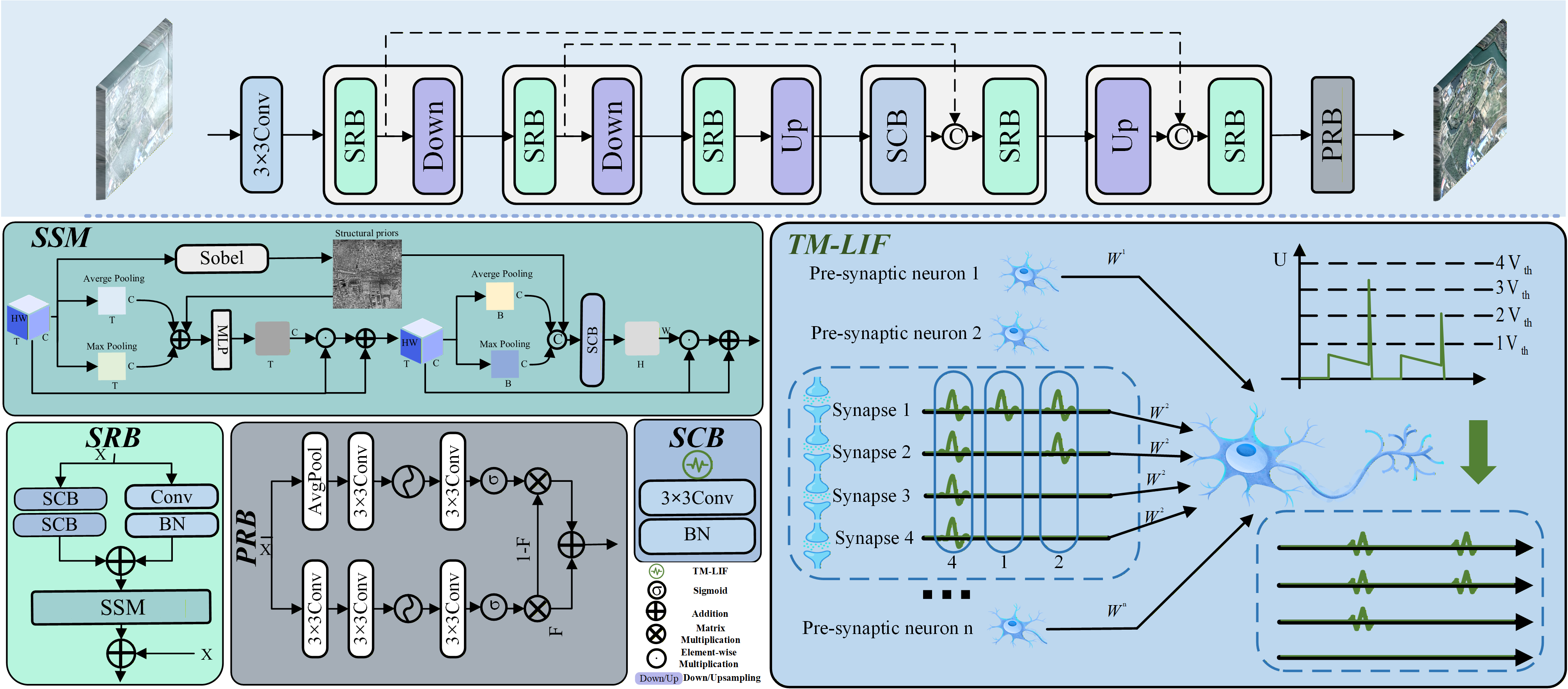}
  \vspace{-0.5em}
  \caption{Overall architecture of EM-SNN. EM-SNN adopts a multi-scale encoder–decoder built from SRB blocks with skip fusion via channel-wise concatenation (C), and uses PRB for reconstruction. TM-LIF is applied in all SCB units across both SRB blocks and the Down/Upsampling stages, while SSM is embedded across scales to provide structure-guided attention.}
  \label{fig:model}
  \vspace{-1.50em}
\end{figure*}

\section{Related Work}

\subsection{Single Image Dehazing}

\noindent\textbf{Prior-based Methods.}
Early single-image dehazing methods estimate atmospheric light and scene transmission using hand-crafted priors, including the Dark Channel Prior (DCP)~\cite{he2010single}, Color Attenuation Prior~\cite{zhu2015fast}, Color-Lines~\cite{fattal2014dehazing}, and Hazelines~\cite{berman2016non}. Remote-sensing-oriented variants further introduce homomorphic filtering and sphere models~\cite{li2018haze}, low-rank and sparse priors~\cite{bi2021haze}, or superpixel-based heterogeneous concentration priors~\cite{liang2024remote}. However, their performance often degrades when the underlying assumptions are violated.

\noindent\textbf{Deep Learning Methods.}
Deep neural networks have become the dominant paradigm for image dehazing. Early CNN models, such as DehazeNet~\cite{cai2016dehazenet} and AOD-Net~\cite{li2017aod}, learn transmission estimation or atmospheric-model inversion, while later methods improve feature representation through attention and multiscale interaction, such as FFA-Net~\cite{qin2020ffa} and GridDehazeNet~\cite{liu2019griddehazenet}. Transformer-based models, including DehazeFormer~\cite{song2023dehazeformer} and Trinity~\cite{chi2023trinity}, further enhance long-range dependency modeling. Recent remote sensing methods, such as SFRDP-Net~\cite{sun2025spatial} and DS-RDMPD~\cite{zhou2025dual}, exploit spatial--frequency fusion or coarse-to-fine diffusion restoration. Despite their strong performance, these models often require substantial computation, whereas lightweight alternatives may sacrifice representation capacity. In contrast, we investigate spiking neural networks for energy-efficient remote sensing image dehazing without compromising restoration quality.

\subsection{Spiking Neural Networks}

Spiking neural networks (SNNs) have attracted increasing attention owing to their event-driven computation and potential advantages in latency and energy efficiency over conventional artificial neural networks~\cite{fang2025spiking,lei2025spike2former}. They have been applied to various vision tasks, including object detection, deraining, and super-resolution~\cite{luo2024integer,song2024learning,xiao2025spiking}. Since spike generation is non-differentiable, directly trained SNNs commonly employ surrogate gradients for backpropagation through time. Advances in membrane-potential optimization and temporal normalization have improved the trainability of deep SNNs~\cite{lee2016training,zheng2021going}, while residual learning and attention mechanisms further enhance their spatial representation capability~\cite{shi2024spikingresformer,lee2025spiking}.

Nevertheless, pixel-level dense restoration with SNNs remains underexplored. Existing studies mainly focus on super-resolution and deraining, while atmospheric-scattering degradation has received limited attention. Previous SNN-based dehazing methods largely rely on spiking-inspired approximations rather than fully spike-driven restoration pipelines~\cite{zhang2024retina,sudevan2026spiking}. Moreover, remote sensing haze is spatially heterogeneous and severely attenuates high-frequency information. The interaction between haze-induced contrast compression and spike discretization can further suppress weak edges and textures, leading to progressive structural-detail loss across network depth.

\section{Method}
\label{sec:method}

\subsection{Overall Architecture}
In this paper, we introduce EM-SNN for remote sensing imagery. As illustrated in Figure \ref{fig:model}, EM-SNN adopts an encoder–decoder architecture. Shallow features extracted from the input image by a $3\times3$ convolution are fed to a multi-scale encoder with Spiking Residual Blocks (SRBs) interleaved with downsampling~\cite{song2024learning}. The proposed Spike Sobel Modulation (SSM) module incorporates Sobel-derived structural priors into each SRB to enhance edges and textures. Throughout the encoder, spiking feature transformations are carried out through convolutional layers followed by Threshold-Modulated leaky integrate-and-fire (TM-LIF) neurons. In the decoder, Up modules with skip connections progressively recover the spatial resolution. Following~\cite{song2024learning}, a Prediction Reconstruction Block (PRB) converts decoded spike features into continuous representations; the final output is obtained by a $3\times3$ convolution.

\subsection{Threshold-Modulated LIF (TM-LIF)}

Remote sensing dehazing is a dense prediction task that requires stable feature magnitudes and consistent activation scales across network depth. However, fixed-threshold LIF neurons are sensitive to haze-induced contrast compression and layer-wise distribution shifts, which may push weak responses below the firing boundary and cause progressive information attenuation and scale inconsistency in deep SNNs~\cite{zhao2025threshold}. Normalized Integer LIF (NI-LIF) alleviates gradient instability by normalizing multi-level spike counts with a fixed quantization factor $D$, thereby constraining spiking outputs to a bounded range~\cite{lei2025spike2former}. Nevertheless, fixed and layer-wise shared quantization boundaries are ill-suited to channel-wise scale heterogeneity, leading to quantization mismatch and imbalanced firing that weaken information transmission and representation in deep SNNs~\cite{kim2020spiking,guo2023rmp,guo2023membrane}. Motivated by these limitations, we propose TM-LIF, which replaces the fixed firing threshold with a statistics-calibrated, channel-wise threshold. By aligning firing boundaries with the characteristic membrane-potential scales of individual channels, TM-LIF reduces scale mismatch and supports high-fidelity image restoration.

\noindent\textbf{Statistics-Calibrated Firing Threshold.}
TM-LIF maintains a channel-wise running estimate of membrane-potential dispersion during training. At training iteration $k$, the running statistic is updated as
\begin{equation}
\bar{\sigma}^{(k)}
=
\mu \bar{\sigma}^{(k-1)}
+
(1-\mu)
\operatorname{Std}_{(b,h,w)}
\left(
\operatorname{sg}\left(U^{(k)}\right)
\right),
\end{equation}
where $U^{(k)}\in\mathbb{R}^{B\times C\times H\times W}$ denotes the pre-spike membrane potential, and $\operatorname{Std}_{(b,h,w)}(\cdot)$ computes the standard deviation over the batch and spatial dimensions, yielding a channel-wise statistic in $\mathbb{R}^{1\times C\times1\times1}$. The momentum coefficient $\mu\in(0,1)$ controls the exponential moving average, while $\operatorname{sg}(\cdot)$ stops gradients through the statistic.

The channel-wise firing threshold is then defined as
\begin{equation}
\boldsymbol{v}_{\mathrm{th}}^{(k)}
=
\alpha\bar{\sigma}^{(k)},
\end{equation}
where $\alpha$ controls the effective quantization interval. During training, the running statistics progressively capture the characteristic activation scales of different channels. Consequently, TM-LIF learns channel-specific quantization boundaries that reduce scale mismatch across layers and channels.

\noindent\textbf{Neuron Dynamics.}
Let $\mathbf{I}_t$ denote the synaptic input at timestep $t$ and $\mathbf{H}_{t-1}\in\mathbb{R}^{B\times C\times H\times W}$ the previous membrane state.
Given the threshold $\boldsymbol{v}_{\mathrm{th}}^{(k)}$, TM-LIF updates the pre-spike membrane potential and produces a multi-level firing output as:
\begin{align}
\mathbf{U}_t &= \beta\,\mathbf{H}_{t-1} + \mathbf{I}_t, \\
\mathbf{S}^{\mathrm{int}}_t &=
\Clip\!\left(
\Round\!\left(\frac{\mathbf{U}_t}{\boldsymbol{v}_{\mathrm{th}}^{(k)}}\right),
\,0,\,D
\right), \\
\mathbf{S}_t &= \frac{1}{D}\,\mathbf{S}^{\mathrm{int}}_t, \\
\mathbf{H}_t &= \mathbf{U}_t - \mathbf{S}^{\mathrm{int}}_t \odot \boldsymbol{v}_{\mathrm{th}}^{(k)}.
\end{align}

Where $\mathbf{U}_t$ integrates the leaked previous state and current input.
$\mathbf{S}^{\mathrm{int}}_t\in\{0,\dots,D\}$ is the integer firing level obtained by quantizing $\mathbf{U}_t$ in units of $\boldsymbol{v}_{\mathrm{th}}^{(k)}$ (as illustrated by the threshold steps in the TM-LIF panel of Fig.~\ref{fig:model}),
while $\mathbf{S}_t\in\{0,\tfrac{1}{D},\dots,1\}$ is the normalized output.
$\beta\in(0,1)$ is the leak factor, and $\Round(\cdot)$ and $\Clip(\cdot,0,D)$ are applied element-wise.

\noindent\textbf{Inference.}
After training, the accumulated running statistics are frozen and used for all test samples. Specifically, the inference-time threshold is given by
\begin{equation}
\boldsymbol{v}_{\mathrm{th}}^{*}
=
\alpha\bar{\sigma}^{*},
\end{equation}
where $\bar{\sigma}^{*}$ denotes the running membrane-dispersion estimate obtained at the end of training. Therefore, the thresholds are channel-specific and conditioned on the training distribution, but remain fixed across test samples. This design avoids batch-dependent statistics and online threshold estimation during inference, leading to deterministic spike quantization without additional adaptation overhead.

\noindent\textbf{Scale Stabilization and Gain Control of TM-LIF.} TM-LIF can be interpreted as a statistics-calibrated re-scaling operation applied before the surrogate nonlinearity. For a mapping $y=f(x)$, local linearization gives $\Delta y\approx J\Delta x$, where $J=\partial y/\partial x$. The spectral norm $\|J\|_2$ bounds the worst-case amplification of local forward perturbations and back-propagated gradients. For clarity, we omit the simulation-step and channel indices in the following analysis. Our objective is to (i) maintain comparable pre-nonlinearity scales across layers and channels and (ii) provide an explicit mechanism for regulating the local gain $\|J\|_2$.

\noindent\textbf{(1) Scale calibration via running statistics.} Consider a single layer with $u=Wx$ and a channel-wise threshold $v_{\mathrm{th}}=\alpha\bar{\sigma}$, where $\bar{\sigma}$ denotes the running estimate of membrane-potential dispersion. Let
\begin{equation}
q=\frac{u}{v_{\mathrm{th}}},\qquad y=\varphi(q),
\end{equation}
where $\varphi(\cdot)$ denotes the normalized multi-level firing mapping used with the surrogate gradient. When $\bar{\sigma}$ closely tracks the characteristic dispersion of $u$, we have
\begin{equation}
\operatorname{Std}(q)=\frac{\operatorname{Std}(u)}{\alpha\bar{\sigma}}\approx\frac{1}{\alpha}.
\end{equation}
Therefore, the input to $\varphi(\cdot)$ remains within a comparable scale range across layers and channels, reducing scale mismatch before spike quantization.

\noindent\textbf{(2) Explicit gain control through a Jacobian bound.} Because gradients are stopped through the running statistic, $v_{\mathrm{th}}$ can be treated as locally constant during back-propagation. By the chain rule,
\begin{equation}
J=\frac{\partial y}{\partial x}\approx\frac{1}{v_{\mathrm{th}}}\operatorname{diag}\left(\varphi'(q)\right)W,
\end{equation}
which yields
\begin{equation}
\|J\|_2\leq\frac{\|W\|_2}{\alpha\bar{\sigma}}B_{\varphi},
\end{equation}
where $B_{\varphi}=\sup_q|\varphi'(q)|$ depends on the surrogate firing function but is independent of $\alpha$. This local bound shows that $\alpha$ and the estimated channel-wise dispersion jointly regulate the effective gain of the surrogate mapping.

\noindent\textbf{(3) Training-time calibration and frozen inference.} During training, the running statistic progressively estimates the characteristic channel-wise dispersion of the membrane potential. At inference, both the accumulated statistic and its corresponding threshold are frozen, providing deterministic channel-wise quantization boundaries calibrated to the training distribution.

\begin{figure}[t]
  \centering
  \includegraphics[width=1.0\linewidth]{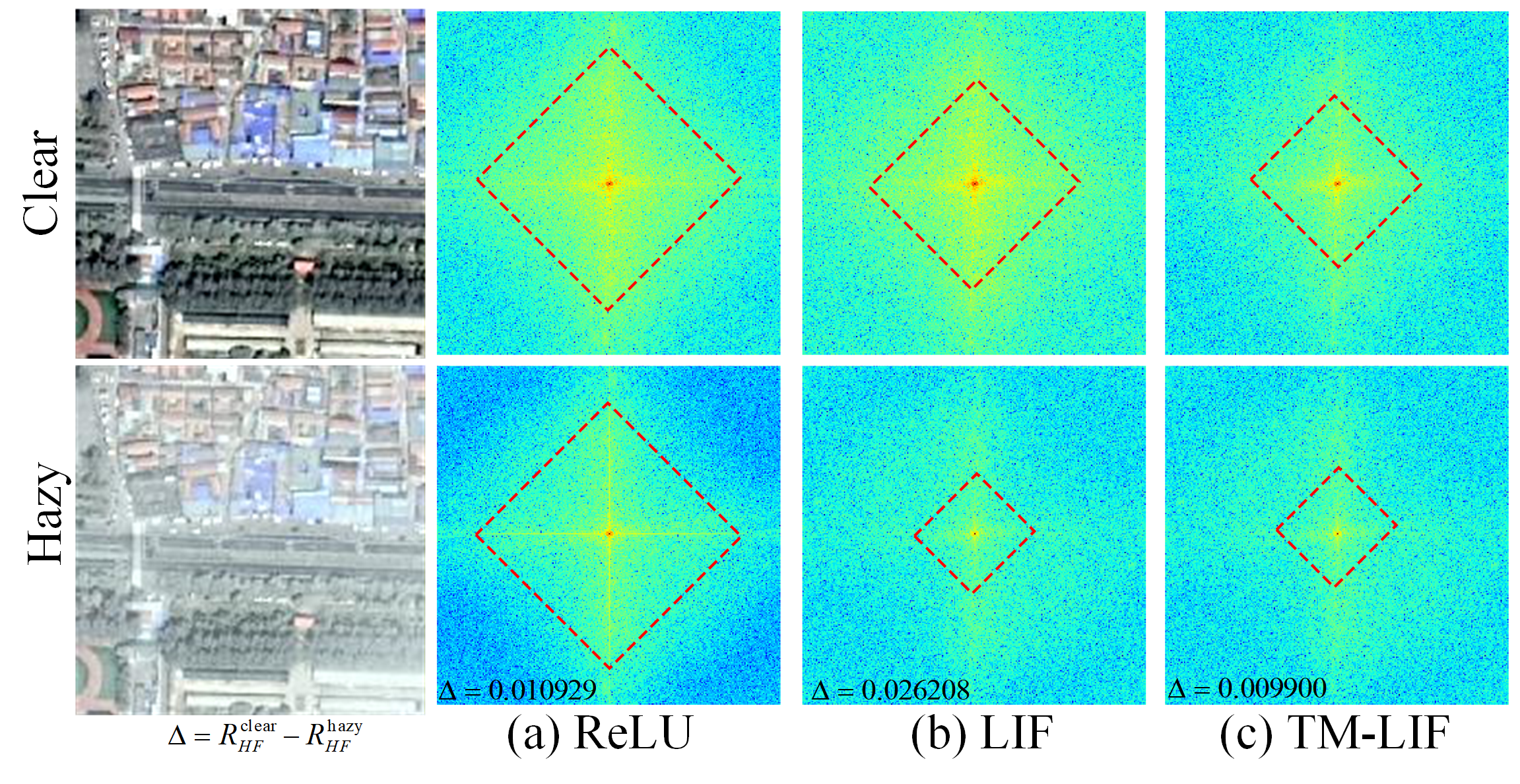}
  \vspace{-1.8em}
  \caption{Frequency-domain visualizations under different neurons. The left shows clear and hazy remote sensing images with the region of interest highlighted, while the right compares ReLU, LIF, and TM-LIF in terms of spectra. We report $\Delta$, the high-frequency energy ratio drop from clear to hazy, which indicates that haze suppresses high-frequency components. Compared with LIF, TM-LIF, via channel-wise adaptive thresholds and multi-level quantization, yields spectra closer to the continuous responses of ReLU and alleviates the haze-induced loss of high-frequency details.}
  \label{fig:diagram}
  \vspace{-1.5em}
\end{figure}

\subsection{Spike Sobel Modulation (SSM)}
\label{sec:SSM}
Haze attenuates high-frequency structural details, so edges and fine textures enter the network as low-amplitude, low-contrast cues (Fig.~\ref{fig:diagram}). In SNNs, such weak cues become small and fast-varying membrane fluctuations under multi-step dynamics and are thus further suppressed by the implicit temporal low-pass filtering of LIF subthreshold integration. Specifically, the discrete-time subthreshold update in Eq.~(3) is a first-order recursion. Applying the $Z$-transform yields
\begin{equation}
U(z)=\beta z^{-1}U(z)+I(z),
\end{equation}
which gives the input-to-membrane transfer function
\begin{equation}
G(z)\triangleq \frac{U(z)}{I(z)}=\frac{1}{1-\beta z^{-1}}.
\end{equation}
This corresponds to a first-order IIR filter with a pole at $z=\beta$; as $\beta\to 1$, stronger low-pass smoothing suppresses high-frequency components, and stacking across layers accumulates into blurred structures and lost fine details.

To compensate for this compounded degradation under strict energy constraints, we propose SSM, which uses a lightweight Sobel-derived structural prior for structure-guided channel and spatial modulation, selectively enhancing edge/texture responses in a spike-efficient manner. SSM is deployment-friendly: Sobel is fixed (parameter-free), and the modulation branch is lightweight (a compact MLP and a shallow spiking convolution), adding minimal overhead while preserving event-driven sparsity.

\begin{table*}[!htbp]
\centering
\renewcommand\arraystretch{1}
\setlength{\tabcolsep}{3pt}
\resizebox{0.95\linewidth}{!}{%
\begin{tabular}{l|c|cc|cc|cc|cc|cc}
\toprule[1pt]
\multirow{2}{*}{\bf Method} &
\multirow{2}{*}{\bf Venue} &
\multicolumn{2}{c|}{\bf LHID} &
\multicolumn{2}{c|}{\bf DHID} &
\multicolumn{2}{c|}{\bf RICE1} &
\multicolumn{2}{c|}{\bf RICE2} &
\multicolumn{2}{c}{\bf Complexity} \\
\cmidrule(lr){3-4}\cmidrule(lr){5-6}\cmidrule(lr){7-8}\cmidrule(lr){9-10}\cmidrule(lr){11-12}
& & PSNR$\uparrow$ & SSIM$\uparrow$
  & PSNR$\uparrow$ & SSIM$\uparrow$
  & PSNR$\uparrow$ & SSIM$\uparrow$
  & PSNR$\uparrow$ & SSIM$\uparrow$
  & Param (M) & Energy (mJ) \\
\midrule
DCP~\cite{he2010single}                     & CVPR'09  & 15.86 & 0.7168 & 19.17 & 0.7561 & 16.03 & 0.6966 & 10.37 & 0.4491 & --      & -- \\
DehazeNet~\cite{cai2016dehazenet}           & TIP'16   & 15.91 & 0.7252 & 21.65 & 0.8389 & 20.94 & 0.8503 & 21.16 & 0.8412 & 0.008   & 2.48 \\
FFA-Net~\cite{qin2020ffa}                   & AAAI'20  & 25.05 & 0.8800 & 27.79 & 0.8870 & 31.78 & 0.9543 & 26.81 & 0.7932 & 4.69    & 1322.64 \\
4KDehazing~\cite{zheng2021ultra}            & CVPR'21  & 25.73 & 0.8406 & 23.44 & 0.8076 & 27.15 & 0.9062 & 26.72 & \underline{0.9140} & 34.55 & 477.94 \\
DCI-Net$^{\dagger}$~\cite{zhang2022dense}   & TGRS'22  & 26.00 & 0.8930 & 28.12 & \textbf{0.9080} & 35.03 & 0.9600 & 32.14 & 0.8600 & 26.51 & 123.56 \\
Dehamer~\cite{guo2022dehamer}               & CVPR'22  & 25.36 & 0.8486 & 26.15 & 0.8821 & 27.41 & 0.9059 & 26.73 & 0.8781 & 132.40  & 274.48 \\
DehazeFormer-b~\cite{song2023dehazeformer}  & TIP'23   & \underline{29.73} & \underline{0.8964} & 24.98 & 0.8734 & 28.46 & 0.9053 & 27.61 & 0.8826 & 2.514 & 118.63 \\
Trinity-Net$^{\dagger}$~\cite{chi2023trinity} & TGRS'23 & 25.76 & 0.8780 & 24.78 & 0.8670 & 27.30 & 0.9000 & 24.71 & 0.7940 & 20.14 & 141.59 \\
PCSFormer$^{\dagger}$~\cite{zhang2024proxy} & TGRS'24  & 24.78 & 0.8375 & 28.68 & 0.8693 & 29.03 & 0.8894 & 26.99 & 0.8973 & 3.73 & 127.24 \\
AU-Net$^{\dagger}$~\cite{du2024dehazing}    & TGRS'24  & 28.44 & 0.893 & 26.42 & 0.8910 & -- & -- & -- & -- & 7.14 & 538.43 \\
SFSNiD~\cite{cong2024semi}                  & CVPR'24  & -- & -- & 28.12 & 0.9030 & 35.93 & \textbf{0.9650} & 34.68 & 0.9080 & 8.35 & 319.05 \\
DS-RDMPD$^{\dagger}$~\cite{zhou2025dual}     & TGRS'25  & 29.39 & 0.8854 & 27.91 & 0.9054 & -- & -- & -- & -- & -- & -- \\
SFRDP-Net$^{\dagger}$~\cite{sun2025spatial} & TGRS'25  & -- & -- & \textbf{28.89} & 0.9070 & \textbf{37.38} & \textbf{0.9650} & \underline{35.45} & 0.9100 & 5.08 & 175.21 \\
\textbf{EM-SNN}                             & Our      & \textbf{30.56} & \textbf{0.9106} & \underline{28.83} & \underline{0.9073} & \underline{35.99} & \underline{0.9630} & \textbf{36.25} & \textbf{0.9406} & 4.81 & 43.62 \\
\bottomrule[1pt]
\end{tabular}%
}
\vspace{-0.5em}
\caption{Quantitative comparison on LHID, DHID, RICE1, and RICE2. \textbf{Bold} and \underline{underlined} indicate the best and second-best results, respectively. $^{\dagger}$ Methods specifically designed for remote sensing dehazing.}
\label{tab:rice_dhid_lhid}
\vspace{-1.5em}
\end{table*}

\noindent\textbf{Sobel-derived Structural Prior.}
We instantiate the structural prior in SSM using fixed Sobel gradients.
For each time step $t$, we first obtain a single-channel map $\mathbf{\hat X}_t$ by averaging $\mathbf{X}_t$ along the channel dimension, and then compute the gradient-strength map:
\begin{equation}
{\mathbf{P}}_t = \big|\text{Sob}_x(\mathbf{\hat X}_t)\big|
+
\big|\text{Sob}_y(\mathbf{\hat X}_t)\big|,
\end{equation}
where $\text{Sob}_x(\cdot)$ and $\text{Sob}_y(\cdot)$ denote fixed Sobel operators that compute the horizontal and vertical image gradients, respectively. To make this cue comparable across inputs and time steps, we further normalize it by its spatial mean:
\begin{equation}
\hat{\mathbf{P}}_t
= \frac{\mathbf{P}_t}
{\operatorname{Mean}_{(h,w)}\!\big(\mathbf{P}_t\big)+\epsilon}.
\end{equation}
where $\hat{\mathbf{P}}_t \in \mathbb{R}^{B\times 1\times H\times W}$ denotes the structural prior that highlights edge- and texture-dominant regions and is leveraged to guide channel and spatial modulation within SSM.

\noindent\textbf{Channel-wise Modulation.}
The channel branch focuses on \emph{what} feature channels should be emphasized or suppressed at each time step.
At time step $t$, we compute three pooled descriptors over spatial dimensions $(h,w)$:

\begin{equation}
\begin{alignedat}{2}
\mathcal{D}^{\mathrm{avg}}_t &\,=\, \mathrm{AvgPool}_{(h,w)}(\mathbf{X}_t), \\
\mathcal{D}^{\mathrm{max}}_t &\,=\, \mathrm{MaxPool}_{(h,w)}(\mathbf{X}_t), \\
\mathcal{D}^{\mathrm{hf}}_t  &\,=\, \mathrm{AvgPool}_{(h,w)}(\mathbf{X}_t \odot \hat{\mathbf{P}}_t),
\end{alignedat}
\end{equation}
where $\mathrm{hf}$ denotes a structural-prior-weighted descriptor, and $\mathcal{D}^{\mathrm{avg}}_t$, $\mathcal{D}^{\mathrm{max}}_t$, and $\mathcal{D}^{\mathrm{hf}}_t \in \mathbb{R}^{B\times C\times 1\times 1}$ are the corresponding channel-wise descriptors. These descriptors are aggregated and fed into a shared MLP to generate the channel gate:
\begin{equation}
\mathcal{G}^{\mathrm{cha}}_t
=
\operatorname{MLP}\!\Big(
\mathcal{D}^{\mathrm{avg}}_t
+
\mathcal{D}^{\mathrm{max}}_t
+
\mathcal{D}^{\mathrm{hf}}_t
\Big),
\end{equation}
where $\mathcal{G}^{\mathrm{cha}}_t \in \mathbb{R}^{B\times C\times 1\times 1}$ and the MLP parameters are shared across time steps. To retain identity initialization and allow both enhancement and suppression, we adopt residual re-weighting:
\begin{equation}
\mathbf{X}^{\mathrm{cha}}_t
=\mathbf{X}_t +\mathbf{X}_t \odot \mathcal{G}^{\mathrm{cha}}_t,
\qquad t=1,\dots,T,
\end{equation}

\noindent\textbf{Spatial-wise Modulation.}
While channel-wise modulation selects informative feature dimensions, the spatial branch focuses on \emph{where} the enhancement should be applied.
At each time step $t$, we form a spatial descriptor by concatenating per-pixel intensity statistics of $\mathbf{X}^{\mathrm{cha}}_t$ with the \emph{normalized} structural prior. Then, this spatial descriptor is fed into a lightweight spiking convolution to predict a spatial gate $\mathcal{G}^{s}_t\in\mathbb{R}^{B\times 1\times H\times W}$, and obtain the Sobel-modulated representations:
\begin{equation}
\mathbf{X}'_t = \mathbf{X}^{\mathrm{cha}}_t + \mathbf{X}^{\mathrm{cha}}_t\odot \mathcal{G}^{s}_t, \qquad t=1,\dots,T,
\end{equation}
The convolution operation is also shared across time steps to reduce parameters.

\begin{figure*}[t]
  \centering
  \includegraphics[width=0.95\linewidth]{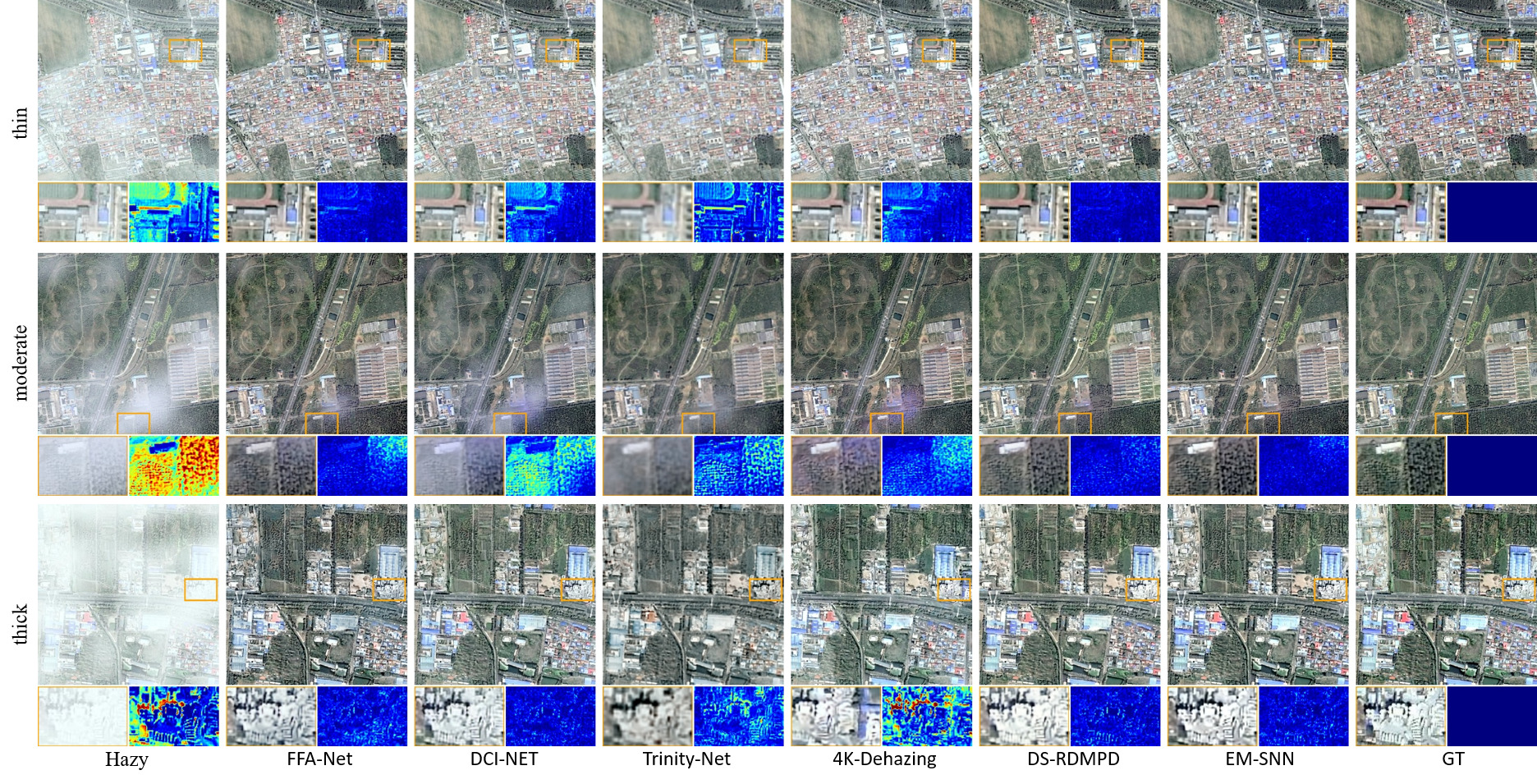}
  \vspace{-1.5em}
  \caption{Qualitative comparisons on SateHaze1K under varying haze densities. The orange box marks the zoom-in region. The bottom rows show enlarged patches and absolute error maps w.r.t. the ground truth (darker blue indicates lower error). Our method recovers clearer structures and sharper boundaries with fewer residual haze artifacts and color distortions.}
  \label{fig:SateHaze1K}
  \vspace{-1.5em}
\end{figure*}

\section{Experiments}

\subsection{Experimental Setup}
\noindent\textbf{Datasets.}
Following prior remote-sensing dehazing benchmarks~\cite{sun2025spatial,zhu2025real}, we evaluate our method on four widely used datasets, including two synthetic datasets (HRSD and SateHaze1K) and two real-world datasets (RICE and RRSHID). HRSD~\cite{zhang2022dense} is a synthetic dataset consisting of LHID and DHID. LHID contains 30,517 training images and 500 test images synthesized from Google Earth imagery, while DHID contains 14,990 images in total, with 14,490 for training and 500 for testing; all images are resized to $512\times512$. SateHaze1K~\cite{huang2020single} contains 1,200 paired hazy/clear GF-2 images categorized into thin, moderate, and thick haze, and each subset is split into 320/35/45 images for training, validation, and testing. For real-world evaluation, RICE~\cite{lin2019remote} consists of RICE1 with 500 Google Earth cloudy/clear pairs and RICE2 with 450 Landsat 8 OLI/TIRS image sets; both are randomly divided into training and testing sets at a ratio of 9:1. RRSHID~\cite{zhu2025real} contains 3,053 paired hazy and haze-free images, which are further divided into thin, moderate, and thick haze subsets with 763, 1,526 and 764 pairs, respectively.

\noindent\textbf{Implementation Details.}

All experiments are implemented in PyTorch and run on an NVIDIA RTX~4090 GPU.
Unless stated otherwise, we set $T{=}1$ and $D{=}4$, with threshold scaling factor $\alpha{=}0.6$, leak factor $\beta{=}0.25$, and EMA momentum $\mu{=}0.9$.
We optimize with AdamW ($\beta_{1}{=}0.9$, $\beta_{2}{=}0.999$) using a cosine-annealing learning-rate schedule, an initial learning rate of $1\times10^{-3}$, and a batch size of 4.
During training, images are randomly cropped to $256\times256$.
The objective function is $\omega_{1}\ell_{1}+\omega_{2}\ell_\mathrm{SSIM}$, with $\omega_{1}{=}1.0$ and $\omega_{2}{=}0.5$, and all models are trained for 1000 epochs.
FLOPs and SOPs are reported at an input resolution of $256\times256$.
The scale-wise depth configuration is $[4,4,8,2,2]$.
For back-propagation through spiking neurons, EM-SNN uses a rectangular surrogate gradient.
Two widely acknowledged metrics, PSNR~\cite{hore2010image} and SSIM~\cite{wang2004image}, are utilized to evaluate all datasets to assess EM-SNN.

\subsection{Experimental Results}

\begin{table}[t]
\centering
\renewcommand\arraystretch{1}
\setlength{\tabcolsep}{3pt}
\resizebox{\linewidth}{!}{
\begin{tabular}{l|cc|cc|cc|cc}
\toprule[1pt]
\multirow{2}{*}{\bf Method} &
\multicolumn{2}{c|}{\bf SateHaze1K-thin} &
\multicolumn{2}{c|}{\bf SateHaze1K-moderate} &
\multicolumn{2}{c|}{\bf SateHaze1K-thick} &
\multicolumn{2}{c}{\bf SateHaze1K-average} \\
\cmidrule(lr){2-3}\cmidrule(lr){4-5}\cmidrule(lr){6-7}\cmidrule(lr){8-9}
& PSNR$\uparrow$ & SSIM$\uparrow$
& PSNR$\uparrow$ & SSIM$\uparrow$
& PSNR$\uparrow$ & SSIM$\uparrow$
& PSNR$\uparrow$ & SSIM$\uparrow$ \\
\midrule
DCP                     & 20.47 & 0.8705 & 20.83 & 0.8971 & 16.71 & 0.7435 & 19.34 & 0.8370 \\
DehazeNet               & 16.82 & 0.8210 & 15.75 & 0.7720 & 15.12 & 0.7430 & 15.90 & 0.7787 \\
FFA-Net                 & 25.16 & 0.9168 & 24.79 & 0.9296 & 22.36 & 0.8453 & 24.10 & 0.8972 \\
4KDehazing              & 23.35 & 0.8939 & 23.73 & 0.9209 & 20.61 & 0.8227 & 22.56 & 0.8792 \\
DCI-Net$^{\dagger}$     & 24.39 & 0.9022 & 25.03 & 0.9315 & 21.82 & 0.8390 & 23.75 & 0.8909 \\
Dehamer                 & 22.50 & 0.8973 & 25.73 & 0.9390 & 17.35 & 0.7957 & 21.86 & 0.8773 \\
DehazeFormer-b          & 25.68 & \underline{0.9193}
                        & 27.15 & 0.9408
                        & 22.74 & 0.8494
                        & 25.19 & 0.9032 \\
Trinity-Net$^{\dagger}$ & 23.46 & 0.8998 & 25.98 & 0.9372 & 20.47 & 0.8148 & 23.30 & 0.8839 \\
PCSFormer$^{\dagger}$   & 22.08 & 0.8751 & 23.29 & 0.9365 & 18.39 & 0.7727 & 21.25 & 0.8614 \\
OK-Net                  & 22.16 & 0.8974 & 26.00 & 0.9437 & 20.51 & 0.8257 & 22.89 & 0.8889 \\
AU-Net$^{\dagger}$      & 22.71 & 0.8992 & 21.04 & 0.9038 & 21.81 & 0.8458 & 21.85 & 0.8829 \\

SFSNiD                  & 25.90 & 0.9180
                        & 26.58 & 0.9370
                        & \textbf{24.35} & \textbf{0.8680}
                        & 25.61 & \underline{0.9077} \\
DS-RDMPD$^{\dagger}$    & \underline{26.03} & 0.9191
                        & \textbf{28.17} & \underline{0.9444}
                        & 22.97 & 0.8485
                        & \textbf{25.72} & 0.9040 \\

\textbf{EM-SNN}         & \textbf{26.29} & \textbf{0.9239}
                        & \underline{27.43} & \textbf{0.9465}
                        & \underline{23.17} & \underline{0.8575}
                        & \underline{25.63} & \textbf{0.9093} \\
\bottomrule[1pt]
\end{tabular}}
\vspace{-1.0em}
\caption{Quantitative comparison on the SateHaze1K.}
\label{tab:satehaze}
\vspace{-1.5em}
\end{table}

\noindent\textbf{Quantitative Results.}
Tables~\ref{tab:rice_dhid_lhid} to ~\ref{tab:rrshid_psnr_ssim_only} report the PSNR and SSIM metrics for all methods.
EM-SNN achieves SOTA performance on the LHID, RRSHID, and RICE2 datasets, compared with other ANN-based methods~\cite{song2023dehazeformer,zheng2021ultra}. Furthermore, EM-SNN consistently surpasses the~\cite{cong2024semi,zhang2022dense}, achieving higher PSNR on both DHID (28.83 dB) and RICE1 (35.99 dB), respectively. 
From the haze-density perspective, EM-SNN performs best under thin haze, achieves the highest SSIM (0.9465) under moderate haze, and ranks second under thick haze in terms of both PSNR (23.17) and SSIM (0.8575), trailing only SFSNiD~\cite{cong2024semi}.
These results demonstrate EM-SNN's robust dehazing quality across datasets, which can be attributed to TM-LIF stabilizing spiking activations via variance-driven threshold modulation under haze-induced contrast compression, and SSM injecting structural priors to strengthen edge/texture recovery.

\noindent\textbf{Qualitative Results.}
To clearly illustrate the dehazing capability of EM-SNN, we present qualitative comparisons on SateHaze1K in Fig.~\ref{fig:SateHaze1K}. The three rows in Fig.~\ref{fig:SateHaze1K} correspond to the thin, moderate, and thick haze levels in SateHaze1K, respectively. Compared with other dehazing methods, EM-SNN restores clearer structures, preserves richer details, and produces fewer artifacts, which is consistent with the quantitative results.

\begin{table}[t]
\centering
\renewcommand{\arraystretch}{1.1}
\setlength{\tabcolsep}{3pt}
\resizebox{\linewidth}{!}{
\begin{tabular}{l|cc|cc|cc|cc}
\toprule
\multirow{2}{*}{Method} &
\multicolumn{2}{c|}{\bf RRSHID-thin} &
\multicolumn{2}{c|}{\bf RRSHID-moderate} &
\multicolumn{2}{c|}{\bf RRSHID-thick} &
\multicolumn{2}{c}{\bf RRSHID-average} \\
\cmidrule(lr){2-3}\cmidrule(lr){4-5}\cmidrule(lr){6-7}\cmidrule(lr){8-9}
& PSNR$\uparrow$ & SSIM$\uparrow$
& PSNR$\uparrow$ & SSIM$\uparrow$
& PSNR$\uparrow$ & SSIM$\uparrow$
& PSNR$\uparrow$ & SSIM$\uparrow$ \\
\midrule
DCP              & 18.46 & 0.4564 & 17.80 & 0.4856 & 18.39 & 0.4843 & 18.22 & 0.4754 \\
FFA-Net          & 17.08 & 0.4452 & 17.40 & 0.5385 & 16.71 & 0.4792 & 17.06 & 0.4876 \\
GridDehazeNet    & 22.77 & 0.6145 & 22.62 & 0.6468 & 23.96 & 0.7112 & 23.12 & 0.6575 \\
4KDehazing       & 22.83 & 0.6177 & 22.47 & 0.6505 & 22.55 & 0.6912 & 22.62 & 0.6531 \\
SCANet           & 18.37 & 0.4718 & 18.11 & 0.5380 & 19.07 & 0.5966 & 18.52 & 0.3741 \\
Trinity-Net      & 20.51 & 0.5728 & 22.46 & 0.5728 & 24.11 & \underline{0.7234} & 22.36 & 0.6230 \\
DehazeFormer-b   & 22.74 & 0.6005 & 23.06 & 0.6137 & 24.69 & 0.7143 & 23.50 & 0.6428 \\
PCSFormer        & 21.83 & 0.5427 & 22.09 & 0.5984 & 23.71 & 0.6547 & 22.54 & 0.5986 \\
PhDNet           & 22.64 & 0.6054 & 22.92 & 0.6448 & 24.28 & 0.6996 & 23.28 & 0.6499 \\
MCAF-Net         & \underline{23.32} & \underline{0.6236} & \underline{23.60} & \underline{0.6583} & \underline{25.40} & 0.7221 & \underline{24.11} & \underline{0.6680} \\
EM-SNN           & \textbf{25.10} & \textbf{0.7286} & \textbf{24.70} & \textbf{0.7115} & \textbf{26.02} & \textbf{0.7682} & \textbf{25.27} & \textbf{0.7361} \\
\bottomrule
\end{tabular}
}
\vspace{-0.5em}
\caption{Performance comparison on RRSHID across haze densities. Results of baseline methods follow~\cite{zhu2025real}.}
\label{tab:rrshid_psnr_ssim_only}
\vspace{-1.0em}
\end{table}

\begin{table}[t]
\centering
\small
\setlength{\tabcolsep}{3.2pt}
\renewcommand{\arraystretch}{1.08}
\resizebox{0.98\columnwidth}{!}{%
\begin{tabular}{@{}lccc@{\hspace{1.4em}}lccc@{}}
\toprule
\textbf{Dataset} & \textbf{FR} & \shortstack{\textbf{SOPs}\\\textbf{(G)}} & \shortstack{\textbf{Energy}\\\textbf{(mJ)}}
& \textbf{Dataset} & \textbf{FR} & \shortstack{\textbf{SOPs}\\\textbf{(G)}} & \shortstack{\textbf{Energy}\\\textbf{(mJ)}} \\
\midrule
SateHaze1K-thin     & 0.1729 & 34.938 & 46.166 & DHID            & 0.2235 & 45.149 & 55.356 \\
SateHaze1K-moderate & 0.1688 & 34.110 & 45.421 & LHID            & 0.1216 & 24.563 & 36.829 \\
SateHaze1K-thick    & 0.1788 & 36.127 & 47.236 & RRSHID-thin     & 0.1450 & 29.298 & 41.090 \\
RICE1               & 0.1626 & 32.845 & 44.283 & RRSHID-moderate & 0.1322 & 26.709 & 38.761 \\
RICE2               & 0.1547 & 31.257 & 42.854 & RRSHID-thick    & 0.1291 & 26.078 & 38.193 \\
\bottomrule
\end{tabular}%
}
\vspace{-0.7em}
\caption{Energy evaluation of EM-SNN across different datasets. FR denotes the average firing rate.}
\label{tab:energy}
\vspace{-1.7em}
\end{table}

\noindent\textbf{Model Efficiency.}
To validate efficiency, we compare the energy consumption of EM-SNN with prior methods. As shown in the last three columns of Tab.~\ref{tab:rice_dhid_lhid}, in view of SNN efficiency is typically assessed in terms of energy, because computation is event-driven and depends on both dense MACs and spike operations (SOPs)~\cite{luo2024integer}.
Following the energy estimation protocol in~\cite{lei2025spike2former}, we report the energy consumption in Tab.~\ref{tab:energy}. Our model maintains low neural activity across datasets, leading to moderate spike operations and low energy consumption.
The average energy consumption is 43.62 mJ. we set the operator energy constant to 4.6 pJ/MAC and use the FLOPs-to-energy-converted upper bound of SFRDP-Net~\cite{sun2025spatial} as the reference. Our method consumes only 25\% of that upper bound.

\subsection{Ablation Study}
For convenience, all ablation experiments are carried out on SateHaze1K and results are averaged over the thin/moderate/thick subsets. Additional experiments and ablation results are provided in the Supplementary Material.

\begin{table}[t]
\centering
\resizebox{0.95\columnwidth}{!}{
\begin{tabular}{c|cc|ccc}
\toprule[1pt]
\bf ID & \bf TM-LIF & \bf SSM & \bf PSNR$\uparrow$ & \bf SSIM$\uparrow$ & \bf Energy (mJ)$\downarrow$ \\
\midrule
(1) &  &   & 24.60 & 0.9019 & 47.76 \\
(2) &\checkmark  &  & 24.93 & 0.9012 & 49.50 \\
(3) &  &\checkmark & 24.80 & 0.9033 & 48.36 \\
(4) & \checkmark & \checkmark & \textbf{25.63} & \textbf{0.9093} & \textbf{46.27} \\
\bottomrule[1pt]
\end{tabular}
}
\vspace{-0.5em}
\caption{Factorial ablation study of two modules on SateHaze1K.}
\label{tab:ablation_factorial_mean}
\vspace{-1.0em}
\end{table}

\begin{table}[t]
\centering
\resizebox{1.0\columnwidth}{!}{
\begin{tabular}{c|l|ccc}
\toprule[1pt]
\bf ID & \bf Variant & \bf PSNR$\uparrow$ & \bf SSIM$\uparrow$ & \bf Energy (mJ)$\downarrow$ \\
\midrule
(1) & LIF + SSM            & 24.80 & 0.9033 & 48.36 \\
(2) & I-LIF + SSM                   & 25.00 & 0.9058 & 46.30 \\
(3) & NI-LIF + SSM                  & 25.00 & 0.9081 & 37.26 \\
(4) & TM-LIF + SSM                  & \textbf{25.63} & \textbf{0.9093} & \textbf{46.27} \\
\midrule
(5) & TM-LIF + w/o SSM              & 24.93 & 0.9012 & 49.50 \\
(6) & TM-LIF + SSM (full)           & \textbf{25.63} & \textbf{0.9093} & \textbf{46.27} \\
(7) & TM-LIF + SSM w/o prior        & 25.30 & 0.9076 & 37.67 \\
(8) & TM-LIF + SSM w/ Scharr prior  & 25.27 & 0.9079 & 38.81 \\
\bottomrule[1pt]
\end{tabular}
}
\vspace{-0.5em}
\caption{Additional ablations on SateHaze1K. The upper group compares neuron designs with SSM fixed, and the lower group compares SSM variants with TM-LIF fixed. Results are averaged over the thin/moderate/thick subsets.}
\label{tab:ablation_statehaze_avg_only}
\vspace{-1.5em}
\end{table}

\noindent\textbf{Impact of Different Designs.}
The baseline is a spiking U-Net with standard LIF neurons ($v_{\mathrm{th}}{=}0.15$). TM-LIF replaces LIF with our threshold-modulated neuron, and SSM is added to the same backbone; Tab.~\ref{tab:ablation_factorial_mean} presents a factorial study of TM-LIF and SSM. Starting from 24.60 dB at 47.76 mJ, adding TM-LIF improves PSNR by +0.33 dB with an energy overhead of +1.73 mJ, while adding SSM yields +0.20 dB and +0.60 mJ. When both are activated, the gain increases to +1.03 dB (SSIM +0.0074) and energy is reduced by 1.49 mJ, indicating complementary effects and a better accuracy–efficiency trade-off.

\begin{figure}[t]
  \centering
  \includegraphics[width=1.0\linewidth]{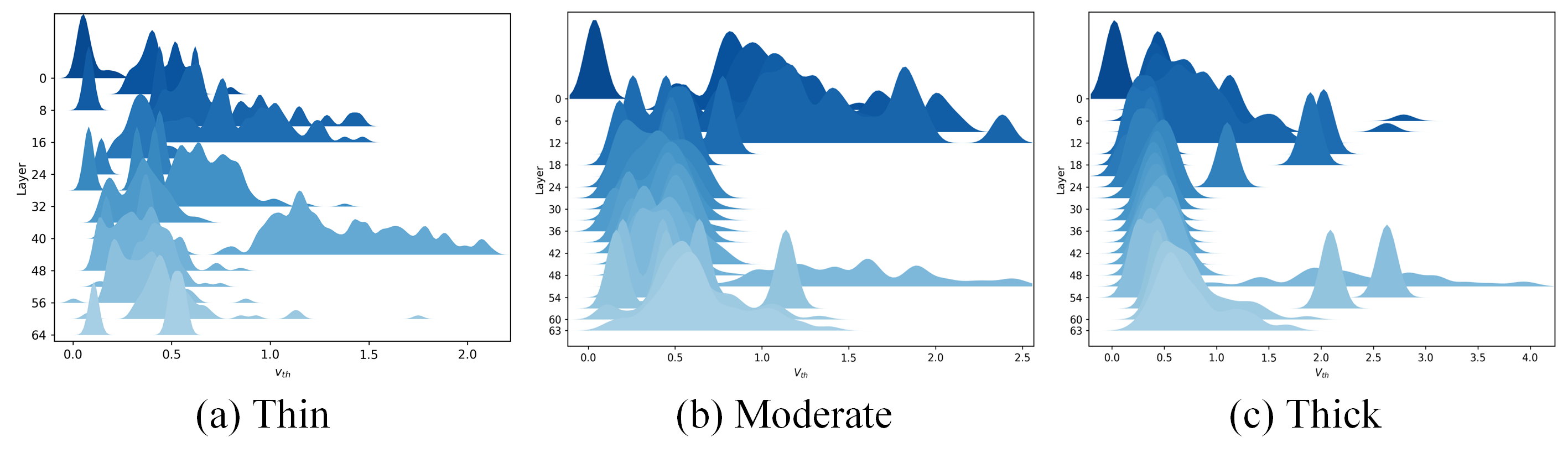}
  \vspace{-1.8em}
  \caption{Layer-wise distributions of frozen channel-wise thresholds $v_{\mathrm{th}}$ learned on (a) thin, (b) moderate, and (c) thick haze subsets (dark-to-light: shallow-to-deep layers).}
\label{fig:vth_dist}
  \vspace{-1.5em}
\end{figure}

Tab.~\ref{tab:ablation_statehaze_avg_only} reports additional ablations on SateHaze1K. In the upper group, different neuron designs are compared with SSM fixed. TM-LIF achieves the best restoration performance, improving PSNR/SSIM from 24.80/0.9033 with standard LIF to 25.63/0.9093, while slightly reducing energy consumption from 48.36 to 46.27 mJ. Compared with TM-LIF, I-LIF~\cite{luo2024integer} exhibits similar energy consumption but lower restoration quality, whereas NI-LIF~\cite{lei2025spike2former} further reduces energy at the cost of a 0.63 dB PSNR drop. These results show that the statistics-calibrated channel-wise thresholds of TM-LIF provide a better accuracy--efficiency balance than fixed-threshold alternatives. In the lower group, different SSM variants are compared with TM-LIF fixed. Removing SSM decreases PSNR/SSIM from 25.63/0.9093 to 24.93/0.9012 and increases energy consumption from 46.27 to 49.50 mJ. Removing the structural prior or replacing Sobel with Scharr reduces energy consumption but also degrades restoration quality. Overall, the full Sobel-based SSM achieves the best restoration performance while maintaining moderate energy consumption.
To visualize how firing thresholds vary with haze density, Figure~\ref{fig:vth_dist} visualizes the frozen channel-wise firing thresholds learned by models separately trained on the thin, moderate, and thick haze subsets. Shallow layers generally exhibit lower and more concentrated thresholds, whereas deeper layers show broader distributions with partially higher values. Models trained on heavier haze exhibit an expanded threshold range and a more pronounced high-threshold tail, indicating that the running statistics capture distribution-dependent shifts in membrane-potential scales rather than sample-wise adaptation during inference.

\begin{table}[t]
\centering
\small
\setlength{\tabcolsep}{3pt}
\renewcommand{\arraystretch}{1.00}
\begin{tabular}{
    cc
    S[table-format=2.2]
    S[table-format=1.4]
    S[table-format=3.2]
}
\toprule
{$T$} & {$D$} & {PSNR $\uparrow$} & {SSIM $\uparrow$} & {Energy (mJ) $\downarrow$} \\
\midrule
1 & 1 & 24.73 & 0.8998 & 30.41 \\
1 & 2 & 25.10 & 0.9076 & 40.04 \\
2 & 2 & 25.37 & 0.9072 & 65.19 \\
\midrule
1 & 4 & 25.63 & 0.9093 & 46.27 \\
2 & 4 & 25.67 & 0.9116 & 80.19 \\
4 & 4 & 25.81 & 0.9135 & 158.37 \\
\midrule
1 & 8 & 25.65 & 0.9092 & 124.91 \\
\bottomrule
\end{tabular}
\vspace{-0.5em}
\caption{Joint ablation of time steps $T$ and spike depth $D$ on SateHaze1K dataset.}
\label{tab:TD_avg}
\vspace{-1.2em}
\end{table}

\noindent\textbf{Impact of Different Time Steps.}
We further study the impact of the number of time steps $T$ and spike depth $D$ on SateHaze1K (Tab.~\ref{tab:TD_avg}). Increasing $T$ or $D$ generally improves reconstruction but raises energy. The lightweight setting $(T{=}1, D{=}1)$ achieves $24.73$ dB/$0.8998$ at $30.41$ mJ, while increasing to $(T{=}1, D{=}4)$ yields $+0.90$ dB/$+0.0095$ at an energy overhead of $+15.86$ mJ, offering a favorable accuracy--energy trade-off. Therefore we adopt $T{=}1,D{=}4$ by default.

\section{Conclusion}
We proposed EM-SNN, a spiking neural network tailored for remote sensing image dehazing. EM-SNN combines a statistics-driven TM-LIF neuron with a SSM module to jointly counter haze-induced contrast compression and the discretization effects of spike generation, thereby improving structural preservation during deep spiking propagation. Experiments on public dehazing datasets demonstrate that EM-SNN achieves competitive restoration quality while maintaining sparse spiking activity and significantly reducing energy consumption. In the future, we will focus on enhancing the robustness of EM-SNN under more diverse real-world sensing conditions and extending its deployment to neuromorphic hardware platforms.

\clearpage
\bibliography{aaai2027}

\end{document}